# Facial Landmark Detection: a Literature Survey

Yue Wu · Qiang Ji

**Abstract** The locations of the fiducial facial landmark points around facial components and facial contour capture the rigid and non-rigid facial deformations due to head movements and facial expressions. They are hence important for various facial analysis tasks. Many facial landmark detection algorithms have been developed to automatically detect those key points over the years, and in this paper, we perform an extensive review of them. We classify the facial landmark detection algorithms into three major categories: holistic methods, Constrained Local Model (CLM) methods, and the regression-based methods. They differ in the ways to utilize the facial appearance and shape information. The holistic methods explicitly build models to represent the global facial appearance and shape information. The CLMs explicitly leverage the global shape model but build the local appearance models. The regression based methods implicitly capture facial shape and appearance information. For algorithms within each category, we discuss their underlying theories as well as their differences. We also compare their performances on both controlled and in the wild benchmark datasets, under varying facial expressions, head poses, and occlusion. Based on the evaluations, we point out their respective strengths and weaknesses. There is also a separate section to review the latest deep learning based algorithms.

The survey also includes a listing of the benchmark databases and existing software. Finally, we identify future research directions, including combining methods in different categories to leverage their respective strengths to solve landmark detection "in-the-wild".

**Keywords** Facial landmark detection · survey

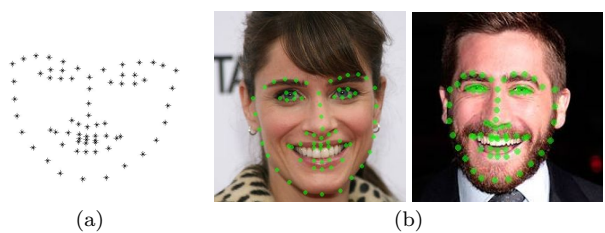

Fig. 1: (a) Facial landmarks defining the face shape. (b) Sample images [10] with annotated facial landmarks.

## 1 Introduction

The face plays an important role in visual communication. By looking at the face, human can automatically extract many nonverbal messages, such as humans' identity, intent, and emotion. In computer vision, to automatically extract those facial information, the localizations of the fiducial facial key points (Figure 1) are usually a key step and many facial analysis methods are built up on the accurate detection of those landmark points. For example, facial expression recognition [68] and head pose estimation algorithms [66] may heavily rely on the facial shape information provided by the landmark locations. The facial landmark points around

Yue Wu
Department of Electrical, Computer, and Systems Engineering, Rensselaer Polytechnic Institute, 110 8th Street, Troy, NY 12180-3590
E-mail: wuyuesophia@gmail.com

Qiang Ji
Department of Electrical, Computer, and Systems Engineering, Rensselaer Polytechnic Institute, 110 8th Street, Troy, NY 12180-3590
E-mail: jiq@rpi.edu

Table 1: Comparison of the major facial landmark detection algorithms.

| Algorithms | Appearance | Shape | Performance | Speed |
|---|---|---|---|---|
| Holistic Method | whole face | explicit | poor generalization/good | slow/fast |
| Constrained Local Method (CLM) | local patch | explicit | good | slow/fast |
| Regression-based Method | local patch/whole face | implicit | good/very good | fast/very fast |

eyes can provide the initial guess of the pupil center positions for eye detection and eye gaze tracking [41]. For facial recognition, the landmark locations on 2D image are usually combined with 3D head model to "frontalize" the face and help reduce the significant within-subject variations to improve recognition accuracy [92]. The facial information gained through the facial landmark locations can provide important information for human and computer interaction, entertainment, security surveillance, and medical applications.

Facial landmark detection algorithms aim to automatically identify the locations of the facial key landmark points on facial images or videos. Those key points are either the dominant points describing the unique location of a facial component (e.g., eye corner) or an interpolated point connecting those dominant points around the facial components and facial contour. Formally, given a facial image denoted as $\mathcal{I}$, a landmark detection algorithm predicts the locations of $D$ landmarks $\mathbf{x} = \{x_1, y_1, x_2, y_2, ..., x_D, y_D\}$, where $x$ and $y$ resentment the image coordinates of the facial landmarks.

Facial landmark detection is challenging for several reasons. First, facial appearance changes significantly across subjects under different facial expressions and head poses. Second, the environmental conditions such as the illumination would affect the appearance of the faces on the facial images. Third, facial occlusion by other objects or self-occlusion due to extreme head poses would lead to incomplete facial appearance information.

Over the past few decades, there have been significant developments of the facial landmark detection algorithms. The early works focus on the less challenging facial images without the aforementioned facial variations. Later, the facial landmark detection algorithms aim to handle several variations within certain categories, and the facial images are usually collected with "controlled" conditions. For example, in "controlled" conditions, the facial poses and facial expressions can only be in certain categories. More recently, the research focuses on the challenging "in-the-wild" conditions, in which facial images can undergo arbitrary facial expressions, head poses, illumination, facial occlusions, etc. In general, there is still a lack of a robust method that can handle all those variations.

Facial landmark detection algorithms can be classified into three major categories: the *holistic methods*, the *Constrained Local Model (CLM) methods*, and *regression-based methods*, depending on how they model the facial appearance and facial shape patterns. The facial appearance refers to the distinctive pixel intensity patterns around the facial landmarks or in the whole face region, while face shape patterns refer to the patterns of the face shapes as defined by the landmark locations and their spatial relationships. As summarized in Table 1, the holistic methods explicitly model the holistic facial appearance and global facial shape patterns. CLMs rely on the explicit local facial appearance and explicit global facial shape patterns. The regression-based methods use holistic or local appearance information and they may embed the global facial shape patterns implicitly for joint landmark detection. In general, the regression-based methods show better performances recently (details will be discussed later). Note that, some recent methods combine the deep learning models and global 3D shape models for detection and they are outside the scope of the three major categories. They will be discussed in detail in Section 4.3.

The remaining parts of the paper are organized as follows. In Sections 2, 3, and 4, we discuss methods in the three major categories: the holistic methods, the Constrained Local Model methods, and the regression-based methods. Section 4.3 is dedicated to the review of the recent deep learning based methods. In Section 5, we discus the relationships among methods in the three major categories. In Section 6, we discuss the limitations of the existing algorithms in "in-the-wild" conditions and some advanced algorithms that are specifically designed to handle those challenges. In Section 7, we discuss related topics, such as face detection, facial landmark tracking, and 3D facial landmark detection. In Section 8, we discuss facial landmark annotations, the popular facial landmark detection databases, software, and the evaluation of the leading algorithms. Finally, we summarize the paper in Section 9, where we point out future directions.

## 2 Holistic methods

Holistic methods explicitly leverage the holistic facial appearance information as well as the global facial shape patterns for facial landmark detection (Figure 2). In the following, we first introduce the classic holistic method: the Active Appearance Model (AAM) [18]. Then, we introduce its several extensions.

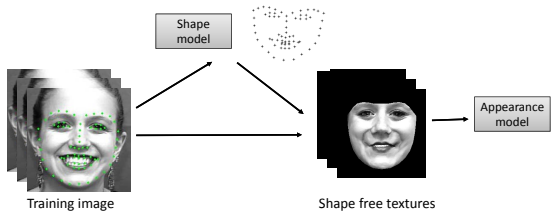

Fig. 2: Holistic Model.

### 2.1 Active Appearance Model

The Active Appearance Model (AAM)[1] was introduced by Taylor and Cootes [27][18]. It is a statistical model that fits the facial images with a small number of coefficients, controlling both the facial appearance and shape variations. During model construction, AAM builds the global facial shape model and the holistic facial appearance model sequentially based on Principal Component Analysis (PCA). During detection, it identifies the landmark locations by fitting the learned appearance and shape models to the testing images

There are a few steps for AAM to construct the appearance and shape models, given the training facial images with landmark annotations, denoted as $\{\mathcal{I}_i, x_i\}_{i=1}^N$, where $N$ is the number of training images. First, Procrustes Analysis [36] is applied to register all the training facial shapes. It removes the affine transformation ($c, \theta, t_c, t_r$ denote the scale, rotation and translation parameters) of each face shape $\mathbf{x}_i$ and generates the normalized training facial shapes $\mathbf{x}'_i$. Second, given the normalized training facial shapes $\{\mathbf{x}'_i\}_{i=1}^N$, PCA is applied to learn the mean shape $\mathbf{s}_0$ and a few orthonormal bases $\{\mathbf{s}_n\}_{n=1}^{K_s}$ that capture the shape variations, where $K_s$ is the number of bases (Figure 3). Given the learned facial shape bases $\{\mathbf{s}_n\}_{n=0}^{K_s}$, a normalized facial shape $\mathbf{x}'$ can be represented using the shape coefficients $\mathbf{p} = \{p_n\}_{n=1}^{K_s}$:

$$\mathbf{x}' = \mathbf{s}_0 + \sum_{n=1}^{K_s} p_n * \mathbf{s}_n. \quad (1)$$

[1] In this paper, we refer Active Appearance Model to the model, independent of the fitting algorithms.

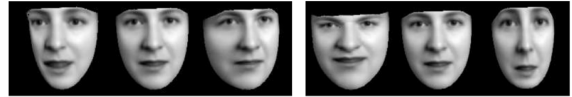

Fig. 3: Learned shape variations using AAM model, adapted from [18].

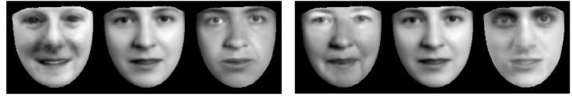

Fig. 4: Learned appearance variations using AAM, adapted from [18].

Third, to learn the appearance model, image wrapping is applied to register the image to the mean shape and generate the shape normalized facial image, denoted as $\mathcal{I}_i(\mathcal{W}(\mathbf{x}'_i))$, where $\mathcal{W}(.)$ indicates the wrapping operation. Then, PCA is applied again on the shape normalized facial images $\{\mathcal{I}_i(\mathcal{W}(\mathbf{x}'_i))\}_{i=1}^N$ to generate a mean appearance $\mathcal{A}_0$ and $K_a$ appearance bases $\mathcal{A} = \{\mathcal{A}_m\}_{m=1}^{K_a}$, as shown in Figure 4. Given the appearance model $\mathcal{A} = \{\mathcal{A}_m\}_{m=0}^{K_a}$, each shape normalized facial image can be represented using the appearance coefficients $\boldsymbol{\lambda} = \{\lambda_m\}_{m=1}^{K_a}$.

$$\mathcal{I}(\mathcal{W}(\mathbf{x}')) = \mathcal{A}_0 + \sum_{m=1}^{K_a} \lambda_m \mathcal{A}_m \quad (2)$$

An optional third model may be applied to learn the correlations among the shape coefficients $\mathbf{p}$ and appearance coefficients $\boldsymbol{\lambda}$.

In landmark detection, AAM finds the shape and appearance coefficients $\mathbf{p}$ and $\boldsymbol{\lambda}$, as well as the affine transformation parameters $\{c, \theta, t_c, t_r\}$ that best fit the testing image, which determine the landmark locations:

$$\mathbf{x} = cR_{2d}(\theta)(\mathbf{s}_0 + \sum_{n=1}^{K_s} p_n * \mathbf{s}_n) + t. \quad (3)$$

Here, $R_{2d}(\theta)$ denotes the rotation matrix, and $t = \{t_c, t_r\}$. To simplify the notation, in the following the shape coefficients would include both PCA coefficients and affine transformation parameters.

In general, the fitting procedure can be formulated by minimizing the distance between the reconstructed images $\mathcal{A}_0 + \sum_{m=1}^{K_a} \lambda_m \mathcal{A}_m$ and the shape normalized testing image $\mathcal{I}(\mathcal{W}(\mathbf{p}))$. The difference is usually referred to as the error image, denoted as $\Delta \mathcal{A}$:

$$\Delta \mathcal{A}(\boldsymbol{\lambda}, \mathbf{p}) = \text{Diff}(\mathcal{A}_0 + \sum_{m=1}^{K_a} \lambda_m \mathcal{A}_m, \mathcal{I}(\mathcal{W}(\mathbf{p}))) \quad (4)$$

$$\boldsymbol{\lambda}^*, \mathbf{p}^* = \arg\min_{\boldsymbol{\lambda}, \mathbf{p}} \Delta\mathcal{A}(\boldsymbol{\lambda}, \mathbf{p}) \quad (5)$$

In the conventional AAM [27][18], model coefficients are estimated by iterative calculation of the error image based on the current model coefficients and model coefficient update prediction based on the error image.

2.2 Fitting algorithms

Most of the holistic methods focus on improving the fitting algorithms, which involves solving Equation 5. They can be classified into the *analytic fitting method* and *learning-based fitting method*.

2.2.1 Analytic fitting methods

The analytic fitting methods formulate AAM fitting problem as a nonlinear optimization problem and solve it analytically. In particular, the algorithm searches the best set of shape and appearance coefficients $\mathbf{p}$, $\lambda$ that minimize the difference between reconstructed image and the testing image with a nonlinear least squares formulation:

$$\tilde{\mathbf{p}}, \tilde{\boldsymbol{\lambda}} = \arg\min_{\mathbf{p}, \boldsymbol{\lambda}} \|\mathcal{A}_0 + \sum_m^M \lambda_m \mathcal{A}_m - \mathcal{I}(\mathcal{W}(\mathbf{p}))\|_2^2. \quad (6)$$

Here, $\mathcal{A}_0 + \sum_m^M \lambda_m \mathcal{A}_m$ represents the reconstructed face in the shape normalized frame depending on the shape and appearance coefficients, and the whole objective function represents the reconstruction error.

One natural way to solve the optimization problem is to use the Gaussian-Newton methods. However, since the Jacobin and Hessian matrix for both $\mathbf{p}$ and $\boldsymbol{\lambda}$ need to be calculated for each iteration [48], the fitting procedure is usually very slow. To address this problem, Baker and Matthews proposed a series of algorithms, among which the Project Out Inverse Compositional algorithm (POIC) [63] and the Simultaneous Inverse Compositional (SIC) algorithm [7] are two popular works. In POIC [63], the errors are projected into space spanned by the appearance eigenvectors $\{\mathcal{A}_m\}_{m=1}^{K_a}$, and its orthogonal complement space. The shape coefficients are firstly searched in the appearance space, and the appearance coefficients are then searched in the orthogonal space, given the shape coefficients. In SIC [7], the appearance and shape coefficients are estimated jointly with gradient descent algorithm. Compared to POIC, SIC is more computationally intensive, but generalizes better than POIC [38].

Recently, more advanced analytic fitting methods only estimate the shape coefficients, which fully determine the landmark locations as in Equation 3. For example, in [4], the Bayesian Active Appearance Model formulates AAM as a probabilistic PCA problem. It treats the texture coefficients as hidden variables and marginalizes them out to solve for the shape coefficients:

$$\tilde{\mathbf{p}} = \arg\max_{\mathbf{p}} ln p(\mathbf{p}) = \arg\max_{p} ln \int_{\boldsymbol{\lambda}} p(\mathbf{p}|\boldsymbol{\lambda}) p(\boldsymbol{\lambda}) d\boldsymbol{\lambda}. \quad (7)$$

But exactly integrating out $\boldsymbol{\lambda}$ can be computationally expensive. To alleviate this problem, in [99], Tzimiropoulos and Pantic propose the fast-SIC algorithm and the fast-forward algorithm. In both algorithms, the appearance coefficient updates $\Delta\boldsymbol{\lambda}$ are represented in terms of the shape coefficient updates $\Delta\mathbf{p}$, and they are plugged into the nonlinear least square formulation, which is then directly minimized to solve for $\Delta\mathbf{p}$. Different from [4], [99] follows a deterministic approach.

2.2.2 Learning-based fitting methods

Instead of directly solving the fitting problem analytically, the learning-based fitting methods learn to predict the shape and appearance coefficients from the image appearances. They can be further classified into *linear regression fitting methods*, *nonlinear regression fitting methods*, and *other learning-based fitting methods*.

**Linear regression fitting methods:** The linear regression fitting methods assume that there is linear relationship between model coefficient updates and the error image $\Delta\mathcal{A}(\boldsymbol{\lambda}, \mathbf{p})$ or image features $\mathcal{I}(\boldsymbol{\lambda}, \mathbf{p})$. They learn linear regression function for the prediction, which follows the conventional AAM as illustrated in the last section.

$$\Delta\mathcal{A}(\boldsymbol{\lambda}, \mathbf{p}) \text{ or } \mathcal{I}(\boldsymbol{\lambda}, \mathbf{p}) \xrightarrow{Linear\ Regression} \Delta\boldsymbol{\lambda}, \Delta\mathbf{p} \quad (8)$$

They therefore estimate the model coefficients by iteratively estimating the model coefficient updates, and add them to the currently estimated coefficients for the prediction in the next iteration. For example, in [26], Canonical Correlation Analysis (CCA) is applied to model the correlation between the error image and the model coefficient updates. It then learns the linear regression function to map the canonical projections of the error image to the coefficient updates. Similarly, in Direct Appearance Model [43], linear model is applied to directly predict the shape coefficient updates from the principal components of the error images. The linear regression fitting methods usually differ in the used image features, linear regression models, and whether to go to a different feature space to learn the mapping [26] [43].

**Nonlinear regression fitting methods:** The linear regression fitting methods assume that the relationship between features and the error image as shown in Equation 4 around the true solution of model coefficients are close to quadratic, which ensures that an iterative procedure with linear updates and adaptive step size would lead to convergence. However, this linear assumption is only true when the initialization is around the true solution which makes the linear regression fitting methods sensitive to the initialization. To tackle this problem, the nonlinear regression fitting methods use nonlinear models to learn the relationship among the image features and the model coefficient updates:

$$\Delta\mathcal{A}(\boldsymbol{\lambda},\mathbf{p}) \; or \; \mathcal{I}(\boldsymbol{\lambda},\mathbf{p}) \xrightarrow{Nonlinear\ Regression} \Delta\boldsymbol{\lambda}, \Delta\mathbf{p} \quad (9)$$

For example, in [83], boosting algorithm is proposed to predict the coefficient updates from the appearance. It combines a set of weak learners to form a strong regressor, and the weak learners are developed based on the Haar-like features and the one-dimensional decision stump. The strong nonlinear regressor can be considered as an additive piecewise function. In [95], Tresadern et al. compared the linear and nonlinear regression algorithms. The used nonlinear regression algorithms include the additive piecewise function developed with boosting algorithm in [83] and the Relevance Vector Machine [104]. They empirically showed that nonlinear regression method is better at the first few iterations to avoid the local minima, while linear regression is better when the estimation is close to the true solution.

#### 2.2.3 Discussion: analytic fitting methods vs. learning-based fitting methods

Compared to the analytic fitting methods solved with gradient descent algorithm with explicit calculation of the Hessian and Jacobian matrices, the learning-based fitting methods use constant linear or nonlinear regression functions to approximate the steepest descent direction. As a result, the learning-based fitting methods are generally fast but they may not be accurate. The analytic methods do not need training images, while the fitting methods do. The learning-based fitting methods usually use a third PCA to learn the joint correlations among the shape and appearance coefficients and further reduce the number of unknown coefficients, while the analytic fitting methods usually do not. But, for the analytic fitting methods, the interaction among appearance and shape coefficients can be embedded in the joint fitting objective function as in Equation 6. The learned correlation between shape and appearance coefficients can reduce the number of parameters. Such learned correlation may not generalize well to different images. The joint estimation of shape and appearance coefficients using Equation 6 can be more accurate. But they are more difficult.

### 2.3 Other extensions

#### 2.3.1 Feature representation

There are other extensions of the conventional AAM methods. One particular direction is to improve the feature representations. It is well known that the AAM model has limited generalization ability and it has difficulty fitting the unseen face variations (e.g. across subjects, illumination, partial occlusion, etc.) [37][38]. This limitation is partially due to the usage of raw pixel intensity as features. To tackle this problem, some algorithms use more robust image features. For example, in [45], instead of using the raw pixel intensity, the wavelet features are used to model the facial appearance. In addition, only the local appearance information is used to improve the robustness to partial occlusion and illumination. In [47], Gabor wavelet with Gaussian Mixture model is used to model the local image appearance, which enables fast local point search. Both methods improve the performances of the conventional AAM method.

#### 2.3.2 Ensemble of AAM models

A single AAM model inherently assumes linearity in face shape and appearance variation. Realizing this limitation, some methods utilize ensemble models to improve the performance. For example, in [71], sequential regression AAM model is proposed, which trains a serials of AAMs for sequential model fitting in a cascaded manner. AAM in the early stage takes into account of the large variations (e.g. pose), while those in the later stage fits to the small variations. In this work, both independent ensemble AAMs and coupled sequential AAMs are used. The independent ensemble AAMs use independently perturbed model coefficients with different settings, while the coupled ensemble AAMs apply the learned prediction model in the first few levels to generate the perturbed training data in the later level. Both boosting regression and random forest regressions are utilized predict the model updates. Similar to the coupled ensemble AAMs in [71], in [82], AAM fitting problem is formulated as an optimization problem with stochastic gradient descent solution. It leads to an approximated algorithm that iteratively learns the linear prediction model from the training data with perturbed model coefficients at the first iteration, and then updates the model coefficients for training data to be used in the next iteration following a cascaded manner. Different from [71], [82] uses different subsets of training data in different stages to escape from the local minima.

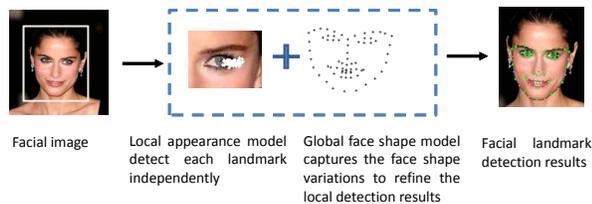

Fig. 5: Constrained Local Method.

## 3 Constrained local methods

As shown in Figure 5, the Constrained Local Model (CLM) methods [24][84] infer the landmark locations **x** based on the global facial shape patterns as well as the independent local appearance information around each landmark, which is easier to capture and more robust to illumination and occlusion, comparing to the holistic appearance.

3.1 Problem formulation

In general, CLM can be formulated either as a deterministic or probabilistic method. In the deterministic view point, CLM [24][84] finds the landmarks by minimizing the misalignment errors subject to the shape patterns:

$$\tilde{\mathbf{x}} = \arg\min_{\mathbf{x}} \mathbf{Q}(\mathbf{x}) + \sum_{d=1}^{D} \mathbf{D}_d(x_d, \mathcal{I}) \qquad (10)$$

Here, $x_d$ represents the positions of different landmarks in **x**. $\mathbf{D}_d(x_d, \mathcal{I})$ represents the local confidence score around $x_d$. $\mathbf{Q}(\mathbf{x})$ represents a regularization term to penalize the infeasible or anti-anthropology face shapes in a global sense. The intuition is that we want to find the best set of landmark locations that have strong independent local support for each landmark and satisfy the global shape constraint.

The shape regularization can be applied to the shape coefficients **p**. If we denote the regularization term as $\mathbf{Q}_p(\mathbf{p})$, Equation 10 becomes:

$$\tilde{\mathbf{p}} = \arg\min_{\mathbf{p}} \mathbf{Q}_p(\mathbf{p}) + \sum_{d=1}^{D} \mathbf{D}_d(x_d(\mathbf{p}), \mathcal{I}) \qquad (11)$$

Here, each landmark location $x_d$ is determined by **p** as in Equation 3.

In the probabilistic view point, CLM can be interpreted as maximizing the product of the prior probability of the facial shape patterns $p(\mathbf{x}; \eta)$ of all points and the local appearance likelihoods $p(x_d|\mathcal{I}; \theta_d)$ of each point:

$$\tilde{\mathbf{x}} = \arg\max_{\mathbf{x}} p(\mathbf{x}; \eta) \prod_{d=1}^{D} p(x_d|\mathcal{I}; \theta_d) \qquad (12)$$

Similar to the deterministic formulation, the prior can also be applied to the shape coefficients **p**, and Equation 12 becomes:

$$\tilde{\mathbf{p}} = \arg\max_{\mathbf{p}} p(\mathbf{p}; \eta_p) \prod_{d=1}^{D} p(x_d(\mathbf{p})|\mathcal{I}; \theta_d). \qquad (13)$$

For both the deterministic and the probabilistic CLMs, there are two major components. The first component is the local appearance model embedded in $\mathbf{D}_d(x_d, \mathcal{I})$ or $p(x_d|\mathcal{I}; \theta_d)$ in Equations 10, 11, 12, and 13. The second component refers to the facial shape pattern constraints either applied to the shape model coefficients **p** or the shape **x** itself, as penalty terms or probabilistic prior distributions. The two components are usually learned separately during training and they are combined to infer landmark locations during landmark detection. In the following, we will discuss each component, and how to combine them for landmark detection.

3.2 Local appearance model

The local appearance model assigns confidence score $\mathbf{D}_d(x_d, \mathcal{I})$ or probability $p(x_d|\mathcal{I}; \theta_d)$ that the landmark with index $d$ is located at a specific pixel location $x_d$ based on the local appearance information around $x_d$ of image $\mathcal{I}$. The local appearance models can be categorized into *classifier-based local appearance model* and the *regression-based local appearance model*.

3.2.1 Classifier-based local appearance model

The classifier-based local appearance model trains binary classifier to distinguish the positive patches that are centered at the ground truth locations and the negative patches that are far away from the ground truth locations. During detection, the classifier can be applied to different pixel locations to generate the confidence scores $\mathbf{D}_d(x_d, \mathcal{I})$ or probabilities $p(x_d|\mathcal{I}; \theta_d)$ through voting. Different image features and classifiers are used. For example, in the original CLM work [24] and FPLL [125], template based method is used to construct the classifier. The original CLM uses the raw image patch, while FPLL uses the HOG feature descriptor. In [11][10], SIFT feature descriptor and SVM classifier are used to learn the appearance model. In [22], Gentle Boost classifier is used. One issue with the classifier-based local appearance model is that it is unclear which feature representation and classifier to use. Even though SIFT and HOG features and SVM classifier are the popular choice, there is some work[105] that learns the features using the deep learning methods, and it is shown that the learned features are comparable to HOG and SIFT.

*3.2.2 Regression-based local appearance model*

During training, the goal of the regression-based local appearance model is to predict the displacement vector $\Delta x_d^* = x_d^* - x$, which is the difference between any pixel location $x$ and the ground truth landmark location $x_d^*$ from the local appearance information around $x$ using the regression models. During detection, the regression model can be applied to patches at different locations $x$ in a region of interest to predict $\Delta x_d$, which can be added to the current location to calculate $x_d$.

$$Regression : \mathcal{I}(x) \to \Delta x_d \qquad (14)$$

$$x_d = x + \Delta x_d; \qquad (15)$$

Predictions from multiple patches can be merged to calculate the final prediction of the confidence score or probability through voting (Figure 6). Different image features and regression models are used. In [22], Cristinacce and Cootes proposed to use the Gentleboost as the regression function. In Cootes's later work [19], he extended the method and used the random forests as the regressor, which shows better performance. In [102], Adaboost feature selection method is combined with SVM regressor to learn the regression function. In [88], nonparametric appearance model is used for location voting based on the contextual features around the landmark. Similar to the classifier-based methods, it's unclear which feature and regression function to use. It is empirically shown in [61] that the LBP features are better than the Haar features and LPQ descriptor. Another issue is that since the regression-based local appearance model performs one-step prediction. The prediction may not be accurate if the current positions are far away from the true target.

*3.2.3 Discussion: local appearance model*

There are several issues related to the local appearance model. First, there exists accuracy-robustness tradeoffs.

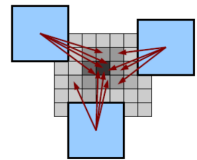

Fig. 6: Regression-based local appearance model [19].

For example, a large local patch is more robust, while it is less accurate for precise landmark localization. A small patch with more distinctive appearance information would lead to more accurate detection results. To tackle this problem, some algorithms [76] combine the large patch and small patch for estimation and adapt the sizes of the patches or searching regions across iterations.

Second, it is unclear which approach to follow, among the classifier-based methods and the regression-based methods. One advantage of the regression-based approach is that it only needs to calculate the features and predict the displacement vectors for a few sample patches in testing. It is more efficient than the classifier-based approach that scans all the pixel locations in the region of interest. It is empirically shown in [22] that the Gentleboost regressor as a regression-based appearance model is better than the Gentleboost classifier as a classifier-based local appearance model.

3.3 Face shape models

The face shape model captures the spatial relationships among facial landmarks, which constrain and refine the landmark location search. In general, they can be classified into *deterministic face shape models* and *probabilistic face shape models*.

*3.3.1 Deterministic face shape models*

The deterministic face shape models utilize deterministic models to capture the face shape patterns. They assign low fitting errors to the feasible face shapes and penalize infeasible face shapes. For example, Active shape model (ASM) [20] is the most popular and conventional face shape model. It learns the linear subspaces of the training face shapes using Principal Component Analysis as in Equation 1. The face can be evaluated by the fitness to the subspaces. It has been used both in the holistic AAM method and the CLM methods. Since one linear ASM may not be effective to model the global face shape variations, in [54], two levels of ASMs are constructed. One level of ASMs are used to capture the shape patterns of each facial component independently, and the other level of ASM is used for modeling the joint spatial relationships among facial components. In [125], Zhu and Ramanan built pose-dependent tree structure face shape model to capture the local nonlinear shape patterns, in which each landmark is represented as a tree node. Improving upon [125], the method in [44] builds two levels of tree structured models focusing on different numbers of landmark points on images with different resolutions. In [9], instead of using the 2D facial shape models, Baltrusaitis et al. proposed to embed the facial shape patterns into a 3D facial deformable model to handle pose variations. During detection, both the 3D model coefficients and the head pose parameters are jointly estimated for landmark detection. This method, however, requires to learn 3D deformable model, and to estimate head pose. 3D landmark detection will be further discussed in Section 7.3.

*3.3.2 Probabilistic face shape models*

The probabilistic face shape models capture the facial shape patterns in a probabilistic manner. They assign high probabilities to face shapes that satisfy the anthropological constraint learned from training data and low probabilities to other infeasible face shapes. The early probabilistic face shape model is the switching model proposed in [94]. It can automatically switch the states of facial components (e.g. mouth open and close) to handle different facial expressions. In [102][61], a generative Boosted Regression and graph Models based method (BoRMaN) constructed based on the Markov Random Field is proposed. Each node in the MRF corresponds to the relative positions of three points, and the MRF as a whole can model the joint relationships among all landmarks. In [11][10], Belhumeur et al. proposed a non-parametric probabilistic face shape model with optimization strategy to fit the facial images. In [108][105], Wu and Ji proposed a discriminative deep face shape model based on the Restricted Boltzmann Machine model. It explicitly handles face pose and expression variation, by decoupling the face shapes into head pose related part and expression related part. Compared to the other probabilistic face shape models, it can better handle facial expressions and poses within a unified model.

*3.3.3 Discussion: face shape model*

Despite of the recent developments of the face shape model, its construction is still an unsolved problem. There is a lack of a unified shape model that can capture all natural facial shape variations (some methods

**Algorithm 1:** Iterative methods

**Data**: The initial searching regions $\Omega^{(1)} = \{\Omega_1^{(1)}, \Omega_2^{(1)}, ..., \Omega_D^{(1)}\}$ for all $D$ landmarks.
**Result**: The detected landmark locations $\mathbf{x}$ or the shape coefficients $\mathbf{p}$ that determine the landmark locations.

1 **for** *t=1 until convergence* **do**
2  Within the searching region $\Omega_d^{(t)}$, detect each facial landmark point independently using the local appearance models, and treat them as the measurements $\mathbf{m}^t$.
3  Refine the measurements $\mathbf{m}$ jointly with the face shape model constraint, and output the estimated locations $\mathbf{x}^t$ or shape coefficients $\mathbf{p}^t$.
4  Modify each searching region $\Omega_d^{(t+1)}$ to be around currently estimated landmark location.

will be discussed in Section 6 ). In addition, it's time-consuming to generate the facial landmark annotations under different facial shape variations, and more complex model requires more data to train. For example, due to facial occlusion, complete landmark annotation is infeasible on the self-occluded profile faces.

3.4 Landmark point detection via optimization

Given the local appearance models and the face shape models described above, CLMs combine them for detection using Equations 10, 11, 12 or 12. This is a non-trivial task, since the analytic representation of the local appearance model is usually not directly computable from the local appearance model and the whole objective function is usually non-convex. To solve this problem, there are two sets of approaches, including the *iterative methods* and the *joint optimization methods*.

*3.4.1 Iterative methods*

The iterative methods decompose the optimization problem into two steps: landmark detection by local appearance model and location refinement by the shape model. They occur alternately until the estimation converges. Specifically, it estimates the optimal landmark positions that best fit the local appearance models for each landmark in the local region independently, and then refines them jointly with the face shape model. The detailed algorithm is shown in Algorithm 1. There are a few algorithms [24][22][19][102][61][108][109] that follow the iterative framework. Those methods differ in the used local appearance models and facial shape models, and we have discussed their particular techniques in the above sections.

*3.4.2 Joint optimization methods*

Different from the iterative methods, the joint optimization methods aim to perform joint inference. The challenge is that they have to find a way to represent the detection results from independent local point detectors and combine them with the face shape model for joint inference. To tackle this problem, in [84], the independent detection results are represented with Kernel Dense Estimation (KDE) and the optimization problem is solved with EM algorithm subject to the shape constraint, which treats the true landmark location as hidden variables. It also discusses some other methods to represent the local detection results, such as using the Isotropic Gaussian Model [20], the Anisotropic Gaussian model [67], and Gaussian Mixture Model [40]. In the Consensus of exemplars work [11][10], in a Bayesian formulation, the local detector is combined with the nonparametric global model. Because of the special probabilistic formulation, the objective function can be optimized in a "brutal-force" way with RANSAC strategy.

There are also algorithms that simplify the shape model to use efficient inference methods. For example, in [125], due to the usage of simple tree structure face shape model, dynamic programming can be applied to solve the optimization problem efficiently. In [23], by converting the objective function into a linear programming problem, Nelder-Meade simplex method is applied to solve the optimization problem.

*3.4.3 Discussion: optimization*

The iterative methods and joint optimization methods have their own benefits and disadvantages. On one hand, the iterative methods are generally more efficient, but it may fail into the local minima due to the iterative procedure. On the other hand, the joint optimization methods are usually more difficult to solve and are computationally expensive.

Note that, as shown in Equations 10, 11, 12 or 12, in CLM, we would either infer the exact landmark locations **x** or the shape coefficients that can fully determine the landmark locations (e.g. shape coefficients **p** when using ASM as the face shape model as in Equation 3). There is a dilemma. On one hand, since small shape model errors may lead to large landmark detection errors, directly predicting the landmark locations may be a better choice. On the other hand, it may be easier to design the cost function for the model coefficients than the shapes. For example, in ASM [20], it is relatively easier to set up the range for the model coefficients while it is difficult to directly set the constraint for the face shapes.

**4 Regression-based methods**

The regression-based methods directly learn the mapping from image appearance to the landmark locations. Different from the Holistic Methods and Constrained Local Model methods, they usually do not explicitly build any global face shape model. Instead, the face shape constraints may be implicitly embedded. In general, the regression-based methods can be classified into *direct regression methods*, *cascaded regression methods*, and *deep-learning based regression methods*. Direct regression methods predict the landmarks in one iteration without any initialization, while the cascaded regression methods perform cascaded prediction and they usually require initial landmark locations. The deep-learning based methods follow either the direct regression or the cascaded regression. Since they use unique deep learning methods, we discuss them separately.

4.1 Direct regression methods

The direct regression methods learn the direct mapping from the image appearance to the facial landmark locations without any initialization of landmark locations. They are typically carried out in one step. They can be further classified into *local approaches* and *global approaches*. The local approaches use image patches, while the global approaches use the holistic facial appearance.

**Local approaches:** The local approaches sample different patches from the face region, and build structured regression models to predict the displacement vectors (target face shape to the locations of the extracted patches), which can be added to the current patch location to produce all landmark locations jointly. The final facial landmark locations can be calculated by merging the prediction results from multiple sampled patches. Note that, this is different from the regression-based local appearance model that predicts each point independently (Section 3.2.2), while the local approaches here predict the updates for all points simultaneously. For example, in [25], conditional Regression Forests are used to learn the mapping from randomly sampled patches in the face region to the face shape updates. In addition, several head pose dependent models are built and they are combined together for detection. Similarly, privileged Information-based conditional random forests model [113] uses additional facial attributes (e.g. head pose, gender, etc.) to train regression forests to predict the face shape updates. Different from [25] which merges the prediction from different pose-dependent models, in testing, it predicts the attributes first and then performs attribute dependent landmark location estimation. In this case, the landmark prediction accuracy will be affected by attribute prediction. One issue with the local regression methods is that the independent local patches may not convey enough information for global shape estimation. In addition, for images with occlusion, the randomly sampled patches may lead to bad estimations.

**Algorithm 2:** Cascaded regression detection
1. Initialize the landmark locations $\mathbf{x}^0$ (e.g. mean face).
2. **for** $t=1, 2, ..., T$ or convergence **do**
3.    Update the landmark locations, given the image and the current landmark location.
   $$f_t : \mathcal{I}, \mathbf{x}^{t-1} \to \Delta\mathbf{x}^t$$
   $$\mathbf{x}^t = \mathbf{x}^{t-1} + \Delta\mathbf{x}^t$$
4. Output the estimated landmark locations $\mathbf{x}^T$.

**Global approaches:** The global approaches learn the mapping from the global facial image to landmark locations directly. Different from the local approaches, the holistic face conveys more information for landmark detection. But, the mapping from the global facial appearance to the landmark locations is more difficult to learn, since the global facial appearance has significant variations, and they are more susceptible to facial occlusion. The leading approaches [91][119] all use the deep learning methods to learn the mapping, which we will discuss in details in Section 4.3. Note that, since the global approaches directly predict landmark locations, they are different from the holistic methods in Section 2 that construct the shape and appearance models and predict the model coefficients.

### 4.2 Cascaded regression methods

In contrast to the direct regression methods that perform one-step prediction, the cascaded regression methods start from an initial guess of the facial landmark locations (e.g. mean face), and they gradually update the landmark locations across stages with different regression functions learned for different stages (Figure 7). Specifically, in training, in each stage, regression models are applied to learn the mapping between shape-indexed image appearances (e.g., local appearance extracted based on the currently estimated landmark locations) to the shape updates. The learned model from the early stage will be used to update the training data for the training in the next stage. During testing, the learned regression models are sequentially applied to update the shapes across iterations. Algorithm 2 summarizes the detection process.

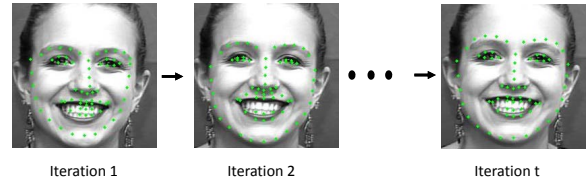

Fig. 7: Cascaded regression methods.

Different shape-indexed image appearance and regression models are used. For example, in [14], Cao et al. proposed the shape-indexed pixel intensity features which are the pixel intensity differences between pairs of pixels whose locations are defined by their relative positions to the current shape. In [76], Ren et al. proposed to learn discriminative binary features by the regression forests for each landmark independently. Then, the binary features from all landmarks are concatenated and a linear regression function is used to learn the joint mapping from appearance to the global shape updates. In [52], ensemble of regression trees are used as regression models for face alignment. By modifying the objective function, the algorithm can use training images with partially labeled facial landmark locations.

Among different cascaded regression methods, the Supervised Descent Method (SDM) in [110] achieves promising performances. It formulates face alignment as a nonlinear least squares problem. In particular, assuming the appearance features (e.g. SIFT) of the local patches around the true landmark locations $\mathbf{x}^*$ are denoted as $\Phi(\mathcal{I}(\mathbf{x}^*))$, the goal of landmark detection is to estimate the location updates $\delta\mathbf{x}$ starting from an initial shape $\mathbf{x}_0$, so that the feature distance is minimized:

$$\begin{aligned}\delta\tilde{\mathbf{x}} &= \arg\min_{\delta\mathbf{x}} f(\mathbf{x}_0 + \delta\mathbf{x}) \\ &= \arg\min_{\delta\mathbf{x}} \|\Phi(\mathcal{I}(\mathbf{x}^*)) - \Phi(\mathcal{I}(\mathbf{x}_0 + \delta\mathbf{x}))\|_2^2\end{aligned} \quad (16)$$

By applying the second order Taylor expansion with Newton-type method, the shape updates are calculated:

$$f(\mathbf{x}_0 + \delta\mathbf{x}) \approx f(\mathbf{x}_0) + \mathbf{J}_f(\mathbf{x}_0)^T \delta\mathbf{x} - \frac{1}{2}\delta\mathbf{x}^T \mathbf{H}_f(\mathbf{x}_0)\delta\mathbf{x} \quad (17)$$

$$\begin{aligned}\delta\mathbf{x} &= -\mathbf{H}_f(\mathbf{x}_0)^{-1}\mathbf{J}_f(\mathbf{x}_0) \\ &= -2\mathbf{H}_f(\mathbf{x}_0)^{-1}\mathbf{J}_\Phi^T(\Phi(\mathcal{I}(\mathbf{x}_0)) - \Phi(\mathcal{I}(\mathbf{x}^*)))\end{aligned} \quad (18)$$

To directly calculate $\delta\mathbf{x}$ analytically is difficult, since it requires the calculation of the Jacobin and Hessian matrix for different $\mathbf{x}_0$. Therefore, supervised descent method is proposed to learn the descent direction with regression method. It is then simplified as the cascaded regression method with linear regression function, which can predict the landmark location updates from shape-indexed local appearance.

$$R = -2\mathbf{H}_f(\mathbf{x}_0)^{-1}\mathbf{J}_\Phi^T \qquad (19)$$

$$b \approx 2\mathbf{H}_f(\mathbf{x}_0)^{-1}\mathbf{J}_\Phi^T \Phi(\mathcal{I}(\mathbf{x}^*)) \qquad (20)$$

$$\delta\mathbf{x} = R\Phi(\mathcal{I}(\mathbf{x}_0)) + b \qquad (21)$$

The regression functions are different for different iterations, but they ignore different possible starting shape $\mathbf{x}_0$ within one iteration.

There are some other variations of the cascaded regression methods. For example, in [6], instead of learning the regression functions in a cascaded manner (the later level depends on the former level), a parallel learning method is proposed, so that the later level only needs the statistic information from the previous level. Based on the parallel learning framework, it's possible to incrementally update the model parameters in each level by adding a few more training samples, which achieves fast training.

The cascaded regression methods are more effective than the direct regression since they follow the coarse-to-fine strategy. The regression functions in the early stage can focus on the large variations while the regression functions in the later stage may focus on the fine search.

However, for cascaded regression methods, it is unclear how to generate the initial landmark locations. The popular choice is to use the mean face, which may be sub-optimal for images with large head poses. To tackle this problem, there are some hybrid methods that use the direct regression methods to generate the initial estimation for cascaded regression methods. For example, in [118], a model based on auto-encoder is proposed. It first performs direct regression on downsampled lower-resolution images, and then refine the prediction in a cascaded manner with higher resolution images. In [123], a coarse to fine searching strategy is employed and the initial face shape is continuously updated based on the estimation from last stage. Therefore, a more close-to-solution initialization will be generated and fine facial landmark detection results are easier to get.

Another issue about the cascaded regression method is that the algorithms apply a fixed number of cascaded prediction, and there is no way to judge the quality of landmark prediction and adapt the necessary cascaded stages for different testing images. In this case, it is possible that the prediction is already trapped in a local minima while the iteration continues. It is also possible that the prediction is already close to the optimum after a few stages. In the existing methods [76], it is only shown that the cascaded regression methods can improve the performance over different cascaded stages, but it doesn't know when to stop.

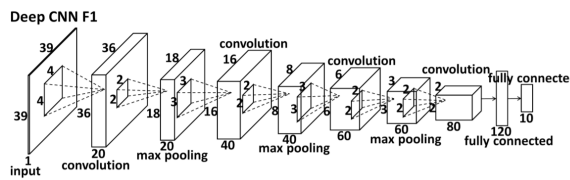

Fig. 8: CNN model structure, adapted from [91].

4.3 Deep learning based methods

Recently, deep learning methods become popular tools for computer vision problems. For facial landmark detection and tracking, there is a trend to shift from traditional methods to deep learning based methods. In the early work [108], the deep boltzmann machine model, which is a probabilistic deep model, was used to capture the facial shape variations due to poses and expressions for facial landmark detection and tracking. More recently, the Convolutional Neural Network (CNN) models become the dominate deep learning models for facial landmark detections, and most of them follow the global direct regression framework or cascaded regression framework. Those methods can be broadly classified into *pure-learning methods* and *hybrid methods*. The pure-learning methods directly predict the facial landmark locations, while the hybrid methods combine deep learning methods with computer vision projection model for prediction.

**Pure-learning methods:** Methods in this category use the powerful CNN models to directly predict the landmark locations from facial images. [91] is the early work and it predicts five facial key points in a cascaded manner. In the first level, it applies a CNN model with four convolution layers (Figure 8) to predict the landmark locations given the facial image determined by the face bounding box. Then, several shallow networks refine each individual point locally.

Ever since then, there are several improvements over [91] in two directions. In the first direction, [119][120] and [75] leverage multi-task learning idea to improve the performance. The intuition is that multiple tasks could share the same representation and their joint relationships would improve the performances of individual tasks. For example, in [119][120], multi-task learning is combined with CNN model to jointly predict facial landmarks, facial head pose, facial attributes etc. A similar multi-task CNN framework is proposed in [75] to jointly perform face detection, landmark localization, pose estimation, and gender recognition. Different from [119] [120], it combines features from multiple convolutional layers to leverage both the coarse and fine feature representations.

In the second direction, some works improve the cascaded procedure of method [91]. For example, in [122], similar cascaded CNN model is constructed to predict many more points (68 landmarks instead of 5). It starts from the prediction of all 68 points and gradually decouples the prediction into local facial components. In [118], the deep auto-encoder model is used to perform the same cascaded landmark search. In [96], instead of training multiple networks in a cascaded manner, Trigeorgis et. al trained a deep convolutional Recurrent Neural Network (RNN) for end-to-end facial landmark detection to mimic the cascaded behavior. The cascaded stage is embedded into the different time slices of RNN.

**Hybrid deep methods:** The hybrid deep methods combine the CNN with 3D vision, such as the projection model and 3D deformable shape model (Figure 9). Instead of directly predicting the 2D facial landmark locations, they predict 3D shape deformable model coefficients and the head poses. Then, the 2D landmark locations can be determined through the computer vision projection model. For example, in [124], a dense 3D face shape model is construct. Then, an iterative cascaded regression framework and deep CNN models are used to update the coefficients of 3D face shape and pose parameters. In each iteration, to incorporate the currently estimated 3D parameters, the 3D shape is projected to 2D using the vision projection model and the 2D shape is used as additional input of the CNN model for regression prediction. Similarly, in [50], in a cascaded manner, the whole facial appearance is used in the first cascaded CNN model to predict the updates of 3D shape parameters and pose, while the local patches are used in the later cascaded CNN models to refine the landmarks.

Compared to the pure-learning methods, the 3D shape deformable model and pose parameters of the hybrid methods are more compact ways to represent the 2D landmark locations. Therefore, there are fewer parameters to estimate in CNN and shape constraint can be explicitly embedded in the prediction. Furthermore, due to the introduction of 3D pose parameters, they can better handle pose variations.

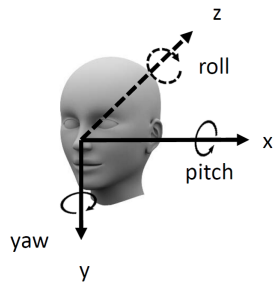

Fig. 9: 3D face model and its projection based on the head pose parameters (i.e. pitch, yaw, roll angles)

Table 2 [2] summarizes the CNN structures of the leading methods. We list their numbers of convolutional layers, the numbers of fully connected layer, whether a 3D model is used, and whether the cascaded method is used. For the cascaded methods, if different model structures are used for different layers, we only list the model in the first level. As can be seen, the models proposed for facial landmark detection usually contain around four convolutional layers and one fully connected layer. The model complexity is on par with the deep models used for other related face analysis tasks, such as head pose estimation [70], age and gender estimation [55], and facial expression recognition [58], which usually have similar or smaller numbers of convolutional layers. For the face recognition problem, the CNN models are usually more complex with more convolutional layers (e.g. eight layers) and fully connected layers [90] [85] [92]. It is in part due to the fact that there is much more training data (e.g., 10M+, 100M+ images) for face recognition comparing to the data set used for facial landmark detections (e.g., 20K+ images) [75]. It is still an open question whether adding more data would improve the performances of facial landmark detection. Another promising direction is to leverage the multi-task learning idea to jointly predict related tasks (e.g., landmark detection, pose, age and gender) with a deeper model to boost the performances for all tasks [75].

4.4 Discussion: regression-based methods

Among different regression methods, cascaded regression method achieves better results than direct regression. Cascaded regression with deep learning can further improve the performance. One issue for the regression-based methods is that since they learn the mapping from the facial appearance within the face bounding box region to the landmarks, they may be sensitive to the used face detector and the quality of the face bounding box. Because the size and location of the initial face is determined by the face bounding box, algorithms trained with one face detector may not work well if a different biased face detector is used in testing. This issue has been studied in [78].

---

[2] For [75], we list the landmark prediction model instead of the multi-task prediction model for fair comparison.

Table 2: CNN model structures of the leading methods.

| methods | # convolutional layer | # fully connected layer, # features | 3D model | cascaded method |
|---|---|---|---|---|
| Sun et. al [91] | 4 | 1 (120) | N | Y |
| Zhang et. al [119] | 4 | 1 (100) | N | N |
| Ranjan et. al [75] | 5 + Dim. reduction | 1 (3072) | N | N |
| Zhou et. al [122] | 4 | 1 (120) | N | Y |
| Zhu et. al [124] | 4 | 2 (256, 234) | Y | Y |
| Jourabloo and Liu [50] | 3 | 1 (150) | Y | Y |

Even though we mentioned that the regression-based methods do not explicitly build the facial shape model, the facial shape patterns are usually implicitly embedded. In particular, since the regression-based methods predict all the facial landmark locations jointly, the structured information as well as the shape constraint are implicitly learned through the process.

## 5 Discussions: relationships among methods in three major categories.

In the previous three sections, we discussed the facial landmark detection methods in three major categories: the holistic methods, the Constrained Local Methods (CLM), and the regression-based methods as summarized in Figure 10. There exist similarities and relationships among the three major approaches.

First, both the holistic methods and CLMs would capture the global facial shape patterns using the explicitly constructed facial shape models, which are usually shared between them. CLMs improve over the holistic methods in that they use the local appearance around landmarks instead of the holistic facial appearance. The motivation is that it's more difficult to model the holistic facial appearances, and the local image patches are more robust to illumination changes and facial occlusion compared to the holistic appearance models.

Second, the regression-based methods, especially for the cascaded regression methods [110] share similar intuitions as the holistic AAM [7][82]. For example, both of them estimate the landmarks by fitting the appearance and they all can be formulated as a nonlinear least squares problem as shown in Equation 6 and 16. However, the holistic methods predict the 2D shape and appearance model coefficients by fitting the holistic appearance model, while the cascaded regression methods predict the landmarks directly by fitting the local appearances without explicit 2D shape model. The fitting problem of holistic methods can be solved with learn-based approaches or analytically as discussed in section 2.2, while all the cascaded regression methods perform estimation by learning. While the learning-based fitting methods for holistic models usually use the same model for coefficient updates in an iterative manner, the cascaded regression methods learn different regression models in a cascaded manner. The AAM model [82] discussed in section 2.3.2 as one particular type of holistic method is very similar to the Supervised Descent Methods (SDM) [110] as one particular type of the cascaded regression method. Both train cascaded models to learn the mapping from shape-indexed features to shape (coefficient) updates. The trained model in the current cascaded stage will modify the training data to train the regression model in the next state. While the former holistic method fits the holistic appearance and predicts the model coefficients, SDM fits the local appearance and predicts the landmark locations.

Third, there are similarities among the regression-based local appearance model used in CLM in section 3.2.2 and the regression-based methods in section 4. Both of them predict the location updates from an initial guess of the landmark locations. The former approach predicts each landmark location independently, while the later approach predicts them jointly, so that shape constraint can be embedded implicitly. The former approach usually performs one-step prediction with the same regression model, while the later approach can apply different regression functions in a cascaded manner.

Fourth, compared to the holistic methods and constrained local methods, the regression-based methods may be more promising. The regression-based methods bypass the explicit face shape modeling and embed the face shape pattern constraint implicitly. The regression-based methods directly predict the landmarks, instead of the model coefficients as in the holistic methods and some CLMs. Directly predicting the shape usually can achieve better accuracy since small model coefficient errors may lead to large landmark errors.

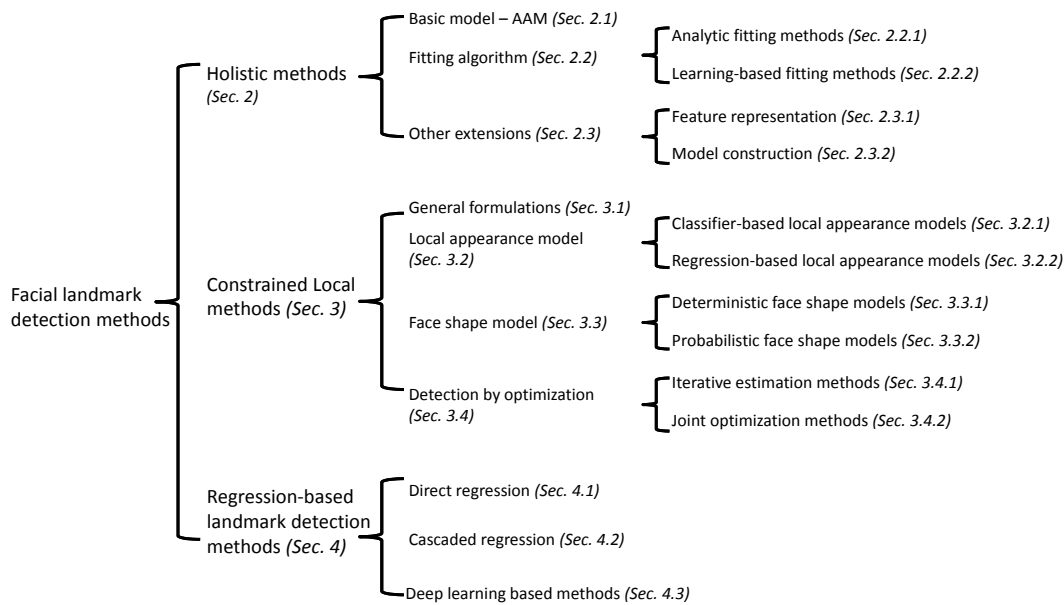

Fig. 10: Major categories of facial landmark detection algorithms

## 6 Facial landmark detection "in-the-wild"

Most of the aforementioned algorithms focus on facial images in "controlled conditions" without significant variations. However, in real world, the facial images would undergo varying facial expressions, head poses, illuminations, facial occlusion etc., which are generally referred to as "in-the-wild" conditions. Some of the aforementioned algorithms may be able to handle those variations implicitly (e.g., deep learning based methods), while some others may fail. In this section, we focus on algorithms that are explicitly designed to handle those challenging conditions.

6.1 Head Poses

Significant head pose (e.g. profile face) is one of the major cause of the failure of the facial landmark detection algorithms (Figure 11). There are a few difficulties. First, the 3D rigid head movement affects the 2D facial appearance and face shape. There would be significant facial appearance and shape variations caused by different head poses. Traditional shape models such as the PCA-based shape model used in AAM and ASM can no longer model the large facial shape variations since they are linear in nature and large facial pose shape variation is non-linear. Second, large head pose may lead to self-occlusion. Due to the missing of the facial landmark points, some facial landmark detection algorithms may not be directly applicable. Third, there is limited training data with large head poses, and it may need extra efforts to annotate the head pose labels to train the algorithms.

To handle large head poses, one direction is to train pose dependent models, and these methods differ in the detection procedures [112][17][125][25]. They either select the best model or merge the results from all models. There are two ways to select the model. The first way is to estimate the head poses using existing head pose estimation methods. For example, in the early work [112], multiple pose dependent AAM models are built in training and the model is selected from the multi-view face detector during testing. In [115], the head pose is first estimated based on the detection of a few facial key points. Then, head pose dependent fitting algorithm is applied to further refine the landmark detection results using the selected pose model. The head pose can also be selected based on the confidence scores using different pose dependent models. For example, in the early work [17], three AAM models are built for faces in different head poses (e.g. left profile, frontal, and right profile) during training. During detection, the result with the smallest fitting error is considered as the final output. In [125], multiple models are built for each discrete head pose and the best fit during testing is outputted as the final result.

The algorithms that select the best head pose dependent model would fail, if the model is not selected correctly. Therefore, it may be better to merge the re-sults from different pose dependent models. For example, in [25], a probabilistic head pose estimator is trained and the facial landmark detection results from different pose dependent models are merged through Bayesian integration.

More recently, there are a few algorithms that build one unified model to handle all head poses. For example, in [106], self-occlusion caused by large head poses is considered as the general facial occlusion, and a unified model is proposed to handle facial occlusion, which explicitly predicts the landmark occlusion along with the landmark locations. In [111], landmark detection follows the cascaded iterative procedure and the pose dependent model is automatically selected based on the estimation from the last iteration. In [124] and [50], Convolutional Neural Network (CNN) is combined with 3D deformable facial shape model to jointly estimate the head pose and facial landmarks on images with large head poses, following the cascaded regression framework. In summary, methods handling head poses include pose dependent shape models, unified pose models, and pose invariant features. They all have their strengths and weaknesses. It depends on the applications to choose which one to follow. Also, for some applications, it may be best to combine different types of methods.

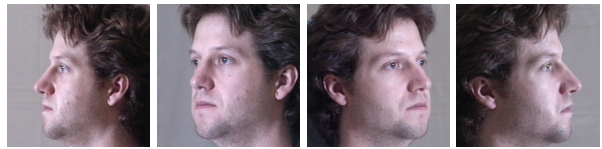

Fig. 11: Facial images [39] with different head poses.

6.2 Occlusion

Facial occlusion is another cause of the failure of the facial landmark detection algorithms. Facial occlusion could be caused by objects or self-occlusion due to large head poses. Figure 12 shows facial images with object occlusion. Some images in Figure 11 also contain facial occlusion. There are a few difficulties to handle facial occlusions. First, the algorithm should rely more on the facial parts without occlusion than the part with occlusion. However, it's difficult to predict which facial part or which facial landmarks are occluded. Second, since arbitrary facial part could be occluded by objects with arbitrary appearances and shapes, the facial landmark detection algorithms should be flexible enough to handle different cases (e.g. mouth may be occluded by mask or nose may be occluded by hand). Third, the occlusion region is usually locally consistent (e.g. it is unlikely that every other point is occluded), but it is difficult to embed this property as a constraint for occlusion prediction and landmark detection.

Due to these difficulties, there is limited work that can handle occlusion. Most of the current algorithms build occlusion dependent models by assuming that some parts of the face are occluded, and merge those models for detection. For example, in [13], face is divided into nine parts and it is assumed that only one part is not occluded. Therefore, the facial appearance information from one particular part is utilized to predict the facial landmark locations and facial occlusions for all parts. The predictions from all nine parts are merged together based on the predicted facial occlusion probability for each part. Similarly, it is assumed in [116] that a few manually pre-designed regions (e.g. one facial component) are occluded, and different models are trained for prediction based on the non-occluded parts.

The aforementioned occlusion dependent models may be sub-optimal, since they assume that a few pre-defined regions are occluded, while the facial occlusion could be arbitrary. Therefore, those algorithms may not cover all rich and complex occlusion cases in real world scenario. Another issue related to the aforementioned methods is that the limited facial appearance information from one part of the face (e.g. from the mouth region) maybe insufficient for the prediction of the facial landmarks in the whole face region. To alleviate this problem, some algorithms handle facial occlusion in a unified framework. For example, in [32], a probabilistic model is proposed to predict the facial landmark locations by joint modeling the local facial appearance around each facial landmark, the landmark visibility for each landmark, occlusion pattern, and hidden states of facial components. It jointly predicts the landmark locations and landmark occlusions through inference in the probabilistic model. In [106], a constrained cascaded regression model is proposed to iteratively predict the facial landmark locations and landmark visibility probabilities, based on local appearance around currently predicted landmark points iteratively. For landmark visibility prediction, it gradually updates the landmark visibility probabilities and it explicitly adds occlusion patterns as a constraint in the prediction. For landmark location prediction, it assigns weights to facial appearance around different facial landmarks based on their landmark visibility probabilities, so that the algorithm relies more on the local appearance from visible landmarks than that from occluded landmarks. Different from [32], the model can handle facial occlusion caused by both object occlusion and large head poses.

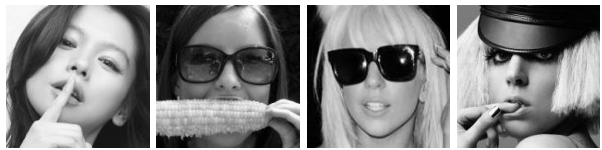

Fig. 12: Facial images [13] with different occlusions.

### 6.3 Facial expression

Facial expression would lead to non-rigid facial motions which would affect facial landmark detection and tracking. For example, as shown in Figure 13, the six basic facial expressions, including happy, surprise, sadness, angry, fear and disgust, would cause the changes of facial appearance and shape. In more natural conditions, facial images would undergo more spontaneous facial expressions other than the six basic expressions. Generally speaking, recent facial landmark detection algorithms have been able to handle facial expressions to some extent.

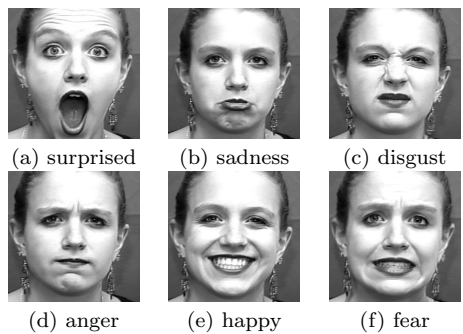

(a) surprised   (b) sadness   (c) disgust
(d) anger       (e) happy     (f) fear

Fig. 13: Facial images [51][59] with different facial expressions.

Even though most algorithms handle facial expressions implicitly, there are some algorithms that are explicitly designed to handle significant facial expression variations. For example, In [94], a hierarchical dynamic probabilistic model is proposed, which can automatically switch between specific states of the facial components caused by different facial expressions. Due to the correlation between facial expression and facial shape, some algorithms also perform joint facial expression and facial landmark detection. For example, in [56], a dynamic bayesian network model is proposed to model the dependencies among facial action units, facial expressions, and face shapes for joint facial behavior analysis and facial landmark tracking. It shows that exploiting the joint relationships and interactions improves the performances of both facial expression recognition and facial landmark detection. Similarly, in [107], a constrained joint cascaded regression framework is proposed for simultaneous facial action unit recognition and facial landmark detection. It first learns the joint relationships among facial action units and facial shapes, and then uses the relationship as a constraint to iteratively update the action unit activation probabilities and landmark locations in a cascaded iterative manner.

There are also some works that handle both facial expression and head pose variations simultaneously for facial landmark detection. For example, In [72][9], 3D face models are used to handle facial expression and pose variations. [9] can predict both the landmarks and the head poses. In [108][105], the face shape model is constructed to handle the variations due to facial expression changes. The model decouples the face shape into expression related parts and head pose related parts. In [121], not only the facial landmark locations, but also the expression and pose are estimated jointly in a cascaded manner with Random Forests model. In addition, in each cascade level, the poses and expressions are firstly updated and they are used further for the estimation of facial landmarks. In [109], a hierarchical probabilistic model is proposed to automatically exploit the relationships among facial expression, head pose, and facial shape changes of different facial components for facial landmark detection. Some multi-task deep learning methods discussed in Section 4.3 also can be included in this category.

## 7 Related topics

### 7.1 Face detection for facial landmark detection

It is usually assumed that face is already detected and given for most of the existing facial landmark detection algorithms. The detected face would provide the initial guess of the face location and face scale. However, there are some issues. First, face detection is still an unsolved problem and it would fail especially on images with large variations. The failure of the face detection would directly lead to the failure of most facial landmark detection algorithms. Second, facial landmark detection accuracy may be significantly affected by the face detectors accuracy. For example, in the regression-based methods, the initial shape is generated by placing the mean shape in the center of the face bounding box, where the scale is also estimated from the bounding box. In CLM, the initial regions of interest for each independent local point detector are determined by the face bounding box. Third, to ensure real-time facial landmark detection, fast face detector is usually preferred.

The most popular face detector has been the Viola-Jones face detector (VJ)[103]. The usage of the integral image and the adaboost learning ensures both fast computation and effective face detection. The part-based approaches [42][29][62] use a slightly different framework. They consider the face as a object consisting of several parts with spatial constraints. More recently, the region-based convolutional neural networks (RCNN) methods [34][33][77] have been used for face detection. They are based on a region proposal component which identifies the possible face regions and a detection component, which further refines the proposal regions for face detection. Generally, the RCNN based methods are more accurate especially for images with large pose, illumination, occlusion variations. However, their computational costs (about 5 frame/second with GPU) are much higher than the traditional face detectors which provide real-time detection. A more detailed study and survey of the face detection algorithms can be found in [117].

There are some algorithms that perform joint face detection and landmark localization. For example, in [125], deformable part model is applied to jointly perform face detection, face alignment and head pose estimation. In [87], face detection and face alignment are formulated as an image retrieval problem. Similarly, In [15], face detection and face alignment are performed jointly in a cascaded regression framework. The face shape is iteratively updated and the bounding box would be rejected if the confidence based on the current face shape is less than a threshold.

7.2 Facial landmark tracking

Facial landmark detection algorithms are generally designed to handle individual facial images, and they can be extended to handle facial image sequences or video. The simplest way is tracking by detection, where facial landmark detection is applied. However, methods in this category ignore the dependency and temporal smoothness among consecutive frames, which are suboptimal.

There are three types of works that perform facial landmark tracking by leveraging the temporal relationship: tracker based independent tracking methods, joint tracking methods, and probabilistic graphical model based methods. The tracker based independent tracking methods [12][94][108] perform facial landmark tracking on the individual points based on the general object trackers, such as the Kalman Filter and Kanade-Lucas-Tomasti tracker (KLT). In each frame, the face shape model is applied to restrict the independently tracked points, so that the face shape in each frame satisfies the face shape pattern constraint [12][94][108]. The joint tracking methods perform facial landmark points update jointly, and they initialize the model parameters or landmark locations based on the information from the last frame. For example, in [3], AAM model coefficients estimated in the last frame are used to initialize the model coefficients in the current frame. In the cascaded regression framework [110], the tracked facial landmark locations in the last frame are used to determine the location and size of the initial facial shape for the cascaded regression in the current frame. Since detection and tracking are initialized differently (detection uses face bounding box to determine the face size and location, while it uses landmark locations in the last frame in tracking), the method needs train different sets of regression functions for detection and tracking. The probabilistic graphical model based methods build dynamic models to jointly embed the spatial and temporal relationships among facial landmark points for facial landmark tracking. For example, Markov Random Field (MRF) model is used in [21], and Dynamic Bayesian Network is used in [56]. The dynamic probabilistic models capture both the temporal coherence as well as shape dependencies among landmark points. More evaluation and review about facial landmark tracking can be found here [16].

7.3 3D facial landmark detection

3D facial landmark detection algorithms detect the 3D facial landmark locations. The existing works can be classified into *2D-data* based methods using 2D images and *3D-data* based methods using 3D face scan data.

Given the limited information, the detection of the 3D facial landmark locations from 2D image is a ill-posed problem. To solve this issue, existing methods either leverage 3D training data or a pre-trained 3D facial shape model and combine them with machine learning. For example, in [97], Tulyakov and Sebe extended the cascaded regression method from 2D to 3D by adding the prediction of the depth information. The method directly learns the regressors to predict the 3D facial landmark locations from 2D image given 3D training data. Since 3D data is difficult to generate, some algorithms learn a 3D facial shape model instead with limited training data. For example, in [35], 2D facial landmarks are firstly detected using the cascaded regression method, and they are combined with a 3D deformable model to determine the face pose and coefficients of the deformable model, based on which they can then recover the positions of the 3D landmark points. Similarly, in [46], cascaded regression method is used to predict both dense 2D facial landmarks and their visibilities. An iterative method is then applied to fit the 2D shape to a pre-trained dense 3D model to estimate the 3D model parameters. There are some methods that use both 3D deformable model and 3D training data. For example, in [49], cascaded regression method is used to estimate the 3D model coefficients and pose parameters from 2D images, which determine both the 2D and the 3D facial landmark locations.

There are a few algorithms that perform 3D facial landmark detection on 3D face scan. For example, in [69], eight 3D facial landmark locations are estimated. The algorithm first uses two 3D local shape descriptors, including the shape index feature and the spin image to generate landmark location candidates for each landmark. Then the final landmark locations are selected by fitting the 3D face model. Similarly, in [57], 17 dominate landmarks are firstly detected on 3D face scan using their particular geometric properties. Then, a 3D template is matched to the testing face to estimate the 3D locations of 20 landmarks. In [69], dense 3D features are estimated from the 3D point cloud generated with depth sensor. In particular, a triangular surface patch descriptor is designed to select and match the training patches to the randomly generated patches from the testing image. Then, the associated 3D face shapes of the training patches are used to vote the 3D shape of the testing image.

Compared to the 2D facial landmark detection, 3D facial landmark detection is still new. There is lack of a large 3D face database with abundant 3D annotations. Compared to largely available 2D images, 3D face scans are difficult to obtain. Labeling 3D facial landmarks on 2D images or 3D face scan are usually more difficult than 2D landmarks.

## 8 Databases and evaluations

8.1 Landmark annotations

Facial landmark annotations refer to the manual annotations of the groundtruth facial landmark locations on facial images. There are usually two types of facial landmarks: the facial key points and interpolated landmarks. The facial key points are the dominant landmarks on face, such as the eye corners, nose tip, mouth corners, etc. They possess unique local appearance/shape patterns. The interpolated landmark points either describe the facial contour or connect the key points (Figure 14). In the early research, only sparse key landmark points are annotated and detected (Figure 14 (a)). Recently, more points are annotated in the new databases (Figure 14 (b)(c)). For example, in BioID, 20 landmarks are annotated, while there are 68 and 194 landmarks annotated in ibug and Heledominantn databases.

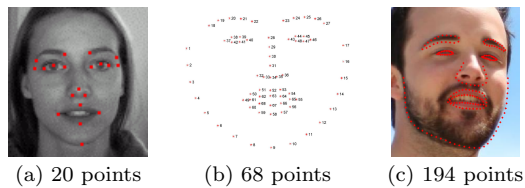

(a) 20 points        (b) 68 points        (c) 194 points

Fig. 14: Facial landmark annotations. Images are adapted from [1][79][54]

There are some issues with the existing landmark annotations. First, the landmark annotations are inherently bias and they are inconsistent across databases. As a result, it's difficult to combine multiple databases for evaluations. The annotation inconsistency also exists for individual landmark annotation. For example, for the annotation of eye corners, some databases tend to provide annotation within the eye region, while the others may annotate the point outside the eye region. To solve this issue, in [89], a method is proposed to combine databases with different facial landmark annotations. It generates a union of landmark annotations by transferring the landmark annotations from source database to target database.

The second issue is that manual annotation is a time-consuming process. There are some works improving the annotation process. In [30], 3D face scan and projection models are used to generate the synthetic 2D facial landmarks and corresponding landmark annotations. The synthetic images are then combined with real images to train the facial landmark detector. In [80], an iterative semi-automatic landmark annotation method is proposed. A facial landmark detector is initially trained with a small number of training data, and it is used to fit new testing images, which are selected by user to retrain the detector. Similarly, in [93], a semi-supervised facial landmark annotation method is proposed. Even though the aforementioned methods improve the facial landmark annotation process, the annotation is still time-consuming and expensive. Overall, the existing training images and databases may still not be adequate for some landmark detection algorithms, such as the deep learning based methods. Finally, to scale up annotation to large datasets, online crowd-sourcing such Amazon Mechanical Turk may be a potential method for facial landmark annotation.

Table 3: Summary of the databases. We use the following notations to represent different variations. e: expression, i: illumination, o: occlusion, p: pose. "(.)" represents moderate/spontaneous variations.

| databases | video(v)/ image(i) | gray(g)/ color(c) | amount of data | variations | number of landmark points |
|---|---|---|---|---|---|
| BioID [1] | i | g | 1521 | e, i | 20 |
| AR [60] | i | c | 4000+ | e, i, o | 22 [2] |
| Extended YaleB [31] | i | g | 16128 | i, p | 3 |
| FERET [74] | i | g | 14501 | e, i, p | 11 [106] |
| CK/CK+ [51][59] | v | g&c | 486 (593) | e | 68 |
| MultiPIE [39] | i | c | 750,000 | e, i, p | 68 or 39 |
| XM2VTSDB [64] | i | c | 1180 | p | 68 [79] |
| FRGC v2[73] | i | c | 50,000 | e, i | 68 [79] |
| BU-4DFE[114] | i | c | 3000 | e | 68 [97] |
| AFLW [53] | i | c | 25,000 | (e, i, o, p) | 21 |
| LFPW [10] | i | c | 1,432 | (e, i, o, p) | 29 (68 [79]) |
| Helen [54] | i | c | 2330 | (e, i, o, p) | 168 (68 [79]) |
| AFW [125] | i | c | 205 | (e, i, o), p | 6 (68 [79]) |
| ibug300 [79] | i | c | 135 | e, i, o, p | 68 |
| ibug300-VW [86] | v | c | 114 | (e, i, o, p) | 68 |
| COFW [13] | i | c | 1852 | (e, i, p), o | 29 |

8.2 Databases

There are two types of databases: databases collected under the "controlled" conditions or databases with "in-the-wild" images. See Table 3 for the summary.

*8.2.1 Databases under "controlled" conditions*

Databases under "controlled" conditions refer to databases with video/images collected indoor with certain restrictions (e.g. pre-defined expressions, head poses etc.).

– BioID [1]: The data set contains 1521 gray scale indoor images with a resolution of 384 × 286 from 23 subjects. Images are taken under different illumination and backgrounds. Subjects may show moderate expression variations. It contains landmark annotations of 20 points.
– AR [60]: The set contains 4,000 frontal color images of 126 people with expressions, illumination, and facial occlusions (e.g. sun glasses and scarf). 22 landmark annotations are provided [2].
– Extended YaleB [31]: The extended Yale Face database B contains 16,128 images of 28 subjects under 9 poses and 64 illumination conditions. The database provides original images, the cropped facial images, and three annotated landmarks.
– FERET [74]: The Facial Recognition Technology (FERET) database contains 14,051 gray scale facial images, covering about 20 discrete head poses that differ in yaw angles. Frontal faces also have illumination and facial expression variations. 11 landmarks on selected profile faces are provided by [106].
– CK/CK+ [51][59]: The Cohn-Kanade AU-coded expression database (CK) contains 486 (593 in CK+) video sequences of frontal faces from 97+ subjects with 6 basic expressions, including happy, surprised, sadness, disgust, fear and anger. The videos starts from the neural expression and goes to apex. CK+ is an extended version of CK database. It includes both posed and spontaneous expressions. AAM landmark tracking results are provided by the database.
– Multi-PIE database [39]: The Multi-PIE face database contains more than 750,000 images of 337 subjects. The facial images are taken under 15 view points and 19 illumination conditions. A few facial expressions are included, such as neutral, smile, surprise, squint, disgust, and scream. 68 or 39 facial landmarks are annotated, depending on the head poses.
– XM2VTSDB [64]: The Extended M2VTS database contains videos of 295 subjects with speech and rotation head movements. 3D head model of each subject is also provided. 68 facial landmark annotations are provided by [79].
– FRGC v2 [73]: The Face Recognition Grand Challenge (FRGC) database contains 50,000 facial images from 4,003 subject sessions with different lighting conditions and two facial expressions (smile and neutral). 3D images acquired by special sensor (Minolta Vivid 900/910) consisting of both range and texture images are also provided. 68 facial landmark annotations on selected images are provided by [79].
– BU-4DFE [114]: The Binghamton University 4D Facial Expression database (BU-4DFE) contains 2D and 3D videos for six prototypic facial expressions (e.g., anger, disgust, happiness, fear, sadness, and surprise) from 101 subjects (58 female and 43 male). There are approximately 60k+ images. 68 2D and

3D facial landmark annotations on selected images are provided by [97]

*8.2.2 "In-the-wild" databases*

Recently, researchers focus on developing more robust and effective algorithms to handle facial landmark detection in real day-life situations. To evaluate the algorithms in those conditions, a few "in-the-wild" databases are collected from the webs, such as Flicks, facebook etc. They contain all sorts of variations, including head pose, facial expression, illumination, ethnicity, occlusion, etc. They are much more difficult than images with "controlled" conditions. Those databases are listed as follows:

- AFLW [53]: The Annotated Facial Landmark in the Wild (AFLW) database contains about 25K images. The annotations include up to 21 landmarks based on their visibility.
- LFPW [10]: The Labeled face parts in the wild (LFPW) database contains 1,432 facial images. Since only the URLs are provided, some images are no longer available. 29 landmark annotations are provided by the original database. Re-annotations of 68 facial landmarks for 1,132 training images and 300 testing images are provided by [79].
- Helen database [54]: The Helen database contains 2,330 high resolution images with dense 194 facial landmark annotations. Re-annotations of 68 landmarks are also provided by [79].
- AFW [125]: The Annotated Faces in the Wild (AFW) database contains about 205 images with relatively larger pose variations than the other "in-the-wild" databases. 6 facial landmark annotations are provided by the database, and re-annotations of 68 landmarks are provided by [79].
- Ibug 300-W [79][81]: The ibug dataset from 300 faces in the Wild (300-W) database [3] is the most challenging database so far with significant variations. It only contains 114 faces of 135 images with annotations of 68 landmarks.
- Ibug 300-VW [86]: The 300 Video in the Wild (300) database contains 114 video sequences for three different scenarios from easy to difficult. 68 facial landmark annotations are provided.
- COFW [13]: The Caltech Occluded Faces in the Wild (COFW) database contains images with significant occlusions. There are 1345 training images and 507 testing images. There are annotations of 29 landmark locations and landmark occlusions.

8.3 Evaluation and discussion

*8.3.1 Evaluation criteria*

Facial landmark detection and tracking algorithms output the facial landmark locations in the facial images or videos. The accuracy is evaluated by comparing the detected landmark locations to the groundtruth facial landmark locations. In particular, if we denote the detected and groundtruth landmark locations for landmark $i$ as $d_i = \{d_{x,i}, d_{y,i}\}$ and $g_i = \{g_{x,i}, g_{y,i}\}$, the detection error for the ith point is:

$$error_i = \|d_i - g_i\|_2 \tag{22}$$

One issue with the above criteria is that the error could change significantly for faces with different sizes. To handle this issue, there are several ways to normalize the error. The *inter-ocular distance* is the most popular criteria. If we denote the left and right pupil centers as $g_{le}$ and $g_{re}$, we can calculate the normalized error as follows:

$$norm\_error_i = \frac{\|d_i - g_i\|_2}{\|g_{le} - g_{re}\|_2} \tag{23}$$

Besides the inter-ocular distance, some works [80] may choose the distance between outer eye corners as a normalization constant. For particular images, such as images with extreme head poses ($\geq 60$ degree) or occlusion (e.g., Figure 11 and 12), the eyes may not be visible. Therefore, some other normalization constants, such as the face size from the face bounding box [125] or the distance between outer eye corner and outer mouth corner (same side of the face) [106] can be used as the normalization constants.

To accumulate the errors of multiple landmarks for one image, the average normalized errors are used:

$$norm\_error\_image = \frac{1}{N} \sum_i^N \frac{\|d_i - g_i\|_2}{\|g_{le} - g_{re}\|_2} \tag{24}$$

To calculate the performances on multiple images, the *mean error* or the *cumulative distribution error* are used. The mean error calculates the mean of the normalized errors of multiple images. The cumulative distribution error calculates the percentages of images that lie under certain thresholds (see Figure 15).

To evaluate the efficiency, the number of processed frames is used. Normally, facial landmark detection algorithms are evaluated on regular PC (e.g., laptop) without powerful GPU or parallel computing implementation etc.

---

[3] Ibug 300-W database contains public available training images and private testing images. The training images include the annotations of public available databases and several newly collected images. Here, we name the newly collected images as Ibug 300-W database

### 8.3.2 Evaluation of existing algorithms

In Table 4, we list the performances of leading algorithms on the benchmark databases, their categories and the landmark detection errors. In Figure 15, we show the cumulative distribution curves of some algorithms on LFPW dtabase. Note that, in this paper, we focus on the reported results from the existing literatures. There are additional detailed references [16] [78] [86] that provide original evaluations by running the software and implementations of known algorithms on different databases.

There are several observations. First, generally, the regression based methods achieve much better performances than the holistic methods and the constrained local model methods, especially on images with significant variations (e.g., ibug 300-w). Second, deep learning based regression methods (e.g., [119]) are the leading techniques and they achieve the state-of-the-art performances on several databases. Third, the performances of the same algorithm are different across database, but the rank of multiple algorithms is generally consistent.

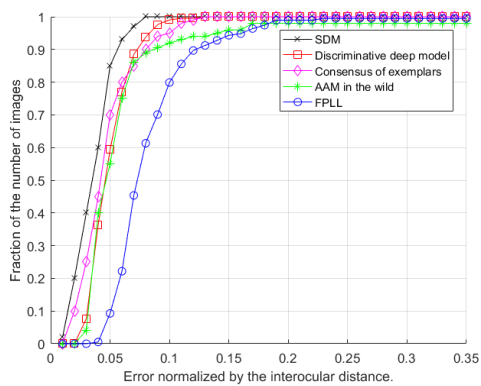

Fig. 15: Comparison of the cumulative distribution curves of several algorithms: SDM [110], Discriminative deep model [105], Consensus of exemplars [10], AAM in the wild [99], and FPLL [125] on LFPW databases [10]. Figure is adapted from [105].

The efficiencies of leading algorithms are shown in Table 5. Note that, the computational speeds of different algorithms are reported from their original papers and their evaluation methods may vary. For example, they have different implementation choices (matlab vs. C++), and they run on different computers. Some algorithms may only report the processing time by excluding the image loading time etc. Generally, we can see that the traditional cascaded regression methods [52][76] are faster than the other methods.

Table 5: Efficiency comparison of leading algorithms. Types: H: holistic methods, C: constrained local methods, R: regression based methods, DR: deep learning based regression methods.

| methods | type | # points | fps |
|---|---|---|---|
| TCDCN [119] | DR | 5 | 58 |
| HyperFace [75] | DR | 21 | 5 |
| Consensus of exemplars [10] | C | 29 | 1 [76] |
| 3DDFA [124] | DR | 68 | 13 |
| CFAN [118] | DR | 68 | 40 |
| CFSS [123] | R | 68 | 25 |
| SDM [110] | R | 68 | 30 |
| 3D Regression [97] | R | 68 | 111 |
| Explicit shape regression [14] | R | 87 | 345 |
| RCPR [13] | R | 194 | 6 |
| One millisecond face alignment [52] | R | 194 | 1000 |
| Face alignment 3000 fps [76] | R | 194 | 200/1500 |

The results shown here are generally consistent with the findings in [16] [78]. In [16], Chrysos et al. show that the one millisecond face alignment [52], the Supervised Descent method [110], and CFSS [123] are good options considering both the speed and accuracy.

### 8.4 Software

In Table 6 and 7, we list a few academic software and commercial software. The academic software refers to the implementations of the existing methods with paper publications. The commercial software is usually only available in a limited sense. For commercial software, visage SDK covers many applications, including facial landmark detection, head pose estimation, and facial expression recognition, which is a good option.

## 9 Conclusion

In this paper, we reviewed the facial landmark detection algorithms in three major categories: the holistic methods, the constrained local methods, and the regression-based methods. In addition, we specifically discussed a few recent algorithms that try to handle facial landmark detection "in-the-wild" under different variations caused by head poses, facial expressions, facial occlusion, strong illumination, low resolution etc. Furthermore, we discussed the popular benchmark databases,

Table 4: Accuracy comparison of leading algorithms. Types: H: holistic methods, C: constrained local methods, R: regression based methods, DR: deep learning based regression methods. The number provided by the original paper is marked as "*". The error normalized by face size is indicated as "fs".

| databases | # points | methods | type | normalized error |
|---|---|---|---|---|
| BU-4DFE [114] | 68[97] | One millisecond face alignment [52] | R | 5.22 [97] |
| | | 3D regression[97] | R | 5.15* |
| AFLW [53] | 21 | Explicit shape regression[14] | R | 8.24 [124] (fs) |
| | | RCPR[13] | R | 7.85 [124] (fs) |
| | | SDM[110] | R | 6.55 [124] (fs) |
| | | 3DDFA [124] | DR | 5.32* (fs) |
| | | HyperFace [75] | DR | 4.26* (fs) |
| AFW[125] | 5 | Explicit shape regression [14] | R | 10.4 [119] |
| | | RCPR[13] | R | 9.3 [119] |
| | | SDM[110] | R | 8.8 [119] |
| | | TCDCN[119] | DR | 8.2* |
| LFPW [10] | 29 | Consensus of exemplars [10] | C | 3.99 [14] |
| | | Robust facial landmark detection [106] | R | 3.93* |
| | | One millisecond face alignment [52] | R | 3.8* |
| | | RCPR[13] | R | 3.50* |
| | | SDM[110] | R | 3.47* |
| | | Explicit shape regression [14] | R | 3.43* |
| | | Face alignment 3000 fps [76] | R | 3.35* |
| | 68 | FPLL[125] | C | 8.29 [123] |
| | | DRMF[5] | C | 6.57 [123] |
| | | RCPR[13] | R | 6.56 [123] |
| | | Gaussian-Newton DPM[100] | C | 5.92 [123] |
| | | SDM[110] | R | 5.67 [123] |
| | | CFAN[118] | DR | 5.44 [123] |
| | | CFSS [123] | R | 4.87* |
| Helen [54] | 68 | FPLL[125] | C | 8.166 [123] |
| | | DRMF[5] | C | 6.70 [123] |
| | | RCPR[13] | R | 5.93 [123] |
| | | Gaussian-Newton DPM [100] | C | 5.69 [123] |
| | | CFAN[118] | DR | 5.53 [123] |
| | | SDM[110] | R | 5.50 [123] |
| | | CFSS[123] | R | 4.63* |
| | 194 | Stasm(ASM)[65] | C | 11.1 [54] |
| | | Component-based ASM[54] | C | 9.1* |
| | | RCPR[13] | R | 6.50* |
| | | SDM[110] | R | 5.85 [76] |
| | | Explicit shape regression [14] | R | 5.7* |
| | | Robust facial landmark detection [106] | R | 5.49* |
| | | Face alignment 3000 fps [76] | R | 5.41* |
| | | One millisecond face alignment [52] | R | 4.9* |
| | | CFSS[123] | R | 4.74* |
| Ibug 300-W [79][81] | 68 | FPLL[125] | C | 10.20 [123] |
| | | DRMF[5] | C | 9.22 [123] |
| | | RCPR[13] | R | 8.35 [123] |
| | | CFAN[118] | DR | 7.69 [28] |
| | | Explicit shape regression [14] | R | 7.58 [76] |
| | | SDM[110] | R | 7.52 [76] |
| | | One millisecond face alignment [52] | R | 6.4 [123] |
| | | Face alignment 3000 fps [76] | R | 6.32* |
| | | 3DDFA [124] | DR | 6.31* |
| | | CFSS[123] | R | 5.76* |
| | | TCDCN[119] | DR | 5.54 [28] |

Table 6: Summary of the academic software

| methods | detection (d) or tracking (t) | realtime (y) or not (n) | source code (sc) or binary code (bc) | number of points | links |
|---|---|---|---|---|---|
| Stasm (ASM) [65] | d | y | sc | 77 | http://www.milbo.users.sonic.net/stasm/ |
| DeMoLib (AAM, ASM etc.) | d | | sc | | http://staff.estem-uc.edu.au/roland/research/demolib-home/ |
| Generic AAM [98] | d | n | sc | 68 | http://ibug.doc.ic.ac.uk/resources/aoms-generic-face-alignment/ |
| AAM in the wild [99] | d | n | sc | 68 | http://ibug.doc.ic.ac.uk/resources/fitting-aams-wild-iccv-2013/ |
| FPLL [125] | d | n | sc | 68 or 39 | http://www.ics.uci.edu/~xzhu/face/ |
| Pose-free [115] | d | n | bc | 66 | http://www.research.rutgers.edu/~xiangyu/face_align.html |
| Flandmark [101] | d | y | sc | 8 | http://cmp.felk.cvut.cz/~uricamic/flandmark/ |
| BoRMaN [102], LEAR [61] | d | n | bc | 20 | http://ibug.doc.ic.ac.uk/resources/facial-point-detector-2010/ |
| DRMF [5] | d | n | bc | 66 | http://ibug.doc.ic.ac.uk/resources/drmf-matlab-code-cvpr-2013/; |
| SDM [110] | d & t | y | bc | 49 | http://www.humansensing.cs.cmu.edu/intraface/index.php |
| Face alignment 3000 fps [76] | d | y | sc | 68 | https://github.com/jwyang/face-alignment |
| RCPR [13] | d&t | y | sc | 29 | http://www.vision.caltech.edu/xpburgos/ICCV13/ |
| One millisecond face alignment [52] | d | y | sc | 194 | http://www.csc.kth.se/~vahidk/face_ert.html |
| CNN [91] | d | y | bc | 5 | http://mmlab.ie.cuhk.edu.hk/archive/CNN_FacePoint.htm |
| Incremental face alignment [6] | d&t | y | bc | 49 | http://ibug.doc.ic.ac.uk/resources/chehra-tracker-cvpr-2014/ |
| Conditional regression forest [25] | d | | sc | 10 | http://www.dantone.me/projects-2/facial-feature-detection/ |
| CLMZ (OpenFace) [9] | d&t | y | sc | 49 | https://github.com/TadasBaltrusaitis/OpenFace/ |
| CCNF [8] | d | y | sc | 30 | https://www.cl.cam.ac.uk/~tb346/res/ccnf.html |

Table 7: Summary of the commercial software. The trial version refers to the free evaluation/download without full access/functionalities.

| methods | detection (d) or tracking (t) | realtime (y) or not (n) | trial version | number of points | links |
|---|---|---|---|---|---|
| Face++ | d | y | y | 83/25/5 | http://www.faceplusplus.com/demo-landmark/ |
| Betaface | d | n | y | 101 | http://betaface.com/wpa/index.php/demo-gallery |
| Lambda Labs | d | y | y | 6 | https://lambdal.com/face-recognition-api#src |
| Visage | d&t | y | y | 51 | http://visagetechnologies.com/products-and-services/visagesdk/ |
| LUXAND | d&t | y | y | 66 | https://www.luxand.com/facesdk/ |

performances of leading algorithms and a few existing software.

There are still a few open questions about facial landmark detection. First, the current facial landmark detection and tracking algorithms still have problems on facial images under challenging "in-the-wild" conditions, including extreme head poses, facial occlusion, strong illumination, etc. The existing algorithms focus on solving one or a few conditions. There is still lack of a facial landmark detection and tracking algorithm that can handle all those cases. Second, there is a lack of a large facial image database that covers all different conditions with facial landmark annotations, which may significantly speed up the development of the algorithms. The existing databases only cover a few conditions (e.g. head poses and expressions). Third, facial landmark detection still heavily relies on the face detection accuracy, which may still fail in certain conditions. Fourth, the computational cost for some landmark detection and tracking algorithms is still high. The facial landmark detection and tracking algorithms should meet the real-time processing requirement.

There are a few future research directions. First, since there are similarities as well as unique properties about the methods in three major approaches, it would be beneficial to have a hybrid approach that combines all three approaches. For example, it would be interesting to see how and whether the appearance and shape models used in holistic methods and CLM can help the regression based methods. It is also interesting to study whether the analytic solutions used for the holistic methods can be applied to solve the cascaded regression at each stage, since they share similar object functions as discussed in section 5. Vice versa, the cascaded regression idea may be applied to the holistic methods to predict the model coefficients in a cascaded manner. Second, currently, the dynamic information is utilized in a limited sense. The facial motion information should be combined with the facial appearance and facial shape for facial landmark tracking. For example, it would be interesting to see how and whether the dynamic features would help facial landmark tracking. Landmark tracking with facial structure information is also an interesting direction. Third, since there are relationships among facial landmark detection and other facial behavior analysis tasks, including head pose estimation and facial expression recognition, their interactions should be utilized for joint analysis. By leveraging their dependencies, we can incorporate the computer vision projection models and improve the performances for all tasks. Finally, to fully exploit the power of deep learning, a large annotated database of millions of images under different conditions is needed. Annotation of such a large image requires a hybrid annotation methods, including human annotation, online crowd sourcing, and automatic annotation algorithms.